\documentclass[10pt,conference]{IEEEtran}
\usepackage{cite}
\usepackage{amsmath,amssymb}
\usepackage{booktabs}
\usepackage{array}
\usepackage{url}
\usepackage{tikz}
\usetikzlibrary{arrows.meta,positioning}

\newcolumntype{L}[1]{>{\raggedright\arraybackslash}p{#1}}
\newcommand{\Pzero}{\textsf{P0}}
\newcommand{\Pone}{\textsf{P1}}
\newcommand{\Ptwo}{\textsf{P2}}

\begin{document}

\title{Separating Automation Gains from Agent Contributions in AI-Assisted Model Validation}

\author{\IEEEauthorblockN{M.~Ahmadi Achachlouei}}

\maketitle

\begin{abstract}
LLM agents are being proposed for bank model validation, yet evaluations that compare agent-assisted work with manual practice credit the agent with gains that come from digitized evidence and deterministic tests. We propose a staged architecture in which an LLM-free controlled workflow (\Pone) is both the fallback and the control arm for individually authorized agent tasks (\Ptwo), and an attribution protocol that credits an agent only for faults \Pone\ does not already report. The protocol charges decoy false alarms, unnecessary actions and endorsed developer assertions, counts invalid outputs as misses, folds newly automatable findings back into \Pone\ with discovery credit, and fixes safety bounds and a break-even cost gate in advance. On a synthetic credit-scorecard harness, five scripted reference policies show why this matters: naive case accuracy (4/6) and pooled detection (3/5) tie four flawed policies, whereas the protocol separates all five and credits only the reference policy with detection beyond \Pone. One fold-back round raises a \Pone-parroting policy's naive scores without any change to the policy. A regression matrix of sixteen mutants covers nine ways in which earlier versions of the evaluation passed misleadingly. No LLM agent has been evaluated yet; we specify a falsifiable pilot.
\end{abstract}


\section{Introduction}

Banks validate the quantitative models behind credit and risk decisions through \emph{effective challenge}: critical, independent analysis by people with the competence and standing to change how a model is used~\cite{sr117,sr262,sudjianto}. Validators reconcile specifications with implementations, probe boundaries, test outcome analyses and write findings. LLM agents can read submissions, call tools and draft findings, and agentic systems for model-risk tasks are already being built~\cite{okpala}. Plausible prose is not assurance, however, and indirect prompt injection turns developer-supplied material into an untrusted input~\cite{greshake,agentdojo}.

In April 2026, SR~26-2 replaced most of SR~11-7~\cite{sr117,sr262}. It excludes generative and agentic AI from its model scope and leaves controls for such tools to banks' own risk management practices, while the models being validated stay in scope~\cite{sr262}. In the EU, creditworthiness assessment is a high-risk AI use~\cite{euaiact}. An AI-enabled \emph{validation workflow}, rather than the model under validation, is therefore an engineering object that banks must govern themselves, and it is our subject.

Two errors motivate this work. First, an agent must never gain the authority to waive tests or approve models. Second, and our focus, comparing agent-assisted validation with manual practice (\Pzero) credits the agent with the gains of digitized evidence, version reconciliation and approved deterministic tests; it can even reward an agent for repeating a discrepancy that those tests already report. Existing agentic model-risk work, for instance, compares its modeling crew with Kaggle solutions and AutoML, and does not compare its model-risk crew with a deterministic validation workflow~\cite{okpala}. Audits of agentic benchmarks document analogous validity failures~\cite{zhu,swebenchplus}, and cost-controlled baselines are essential~\cite{kapoor}.

\textbf{Contributions.} (1)~A staged architecture (Sec.~\ref{sec:arch}) that keeps an LLM-free workflow, \Pone, operational as both the fallback and the control arm for individually authorized agent tasks, \Ptwo. (2)~An attribution protocol (Sec.~\ref{sec:protocol}) that credits an agent only for what \Pone\ does not report, folds automatable findings back into \Pone\ with discovery credit, charges decoys, unnecessary actions and endorsed assertions, scores missing or invalid outputs as misses, and fixes safety and economic gates in advance. (3)~A releasable synthetic harness with evidence about the measurement itself (Sec.~\ref{sec:harness}): scripted policies that the protocol separates and naive metrics do not, one executed fold-back round, and a regression matrix for nine ways the evaluation once passed misleadingly. We claim no individual control as novel and report no agent result; whether any agent adds anything is the question the protocol is built to answer.

\section{Background and Related Work}

\emph{Model validation.} Supervisory guidance and practice guides organize validation around conceptual soundness, implementation testing, outcome analysis and ongoing monitoring. It is performed by a second line of defense that is independent of the model developers (the first line) and is itself checked by internal audit (the third line)~\cite{sr117,sr262,sudjianto}. The harness in Sec.~\ref{sec:harness} seeds faults in implementation, test coverage, outcome analysis and a feature study.

\emph{Validity of agent evaluation.} Absolute-success benchmarks such as AgentBench and $\tau$-bench score agents on fixed tasks~\cite{agentbench,taubench}. Audits of agentic benchmarks report reward-design errors and solution leakage~\cite{zhu,swebenchplus}, Kapoor et al.\ argue for cost-controlled baselines~\cite{kapoor}, and automated judges do not establish factual support~\cite{shankar}. We add a domain control arm: the deterministic workflow a bank would actually operate. Our negative controls draw on the oracle problem, mutation testing and common-mode failures of diverse implementations~\cite{barr,jia,knight}.

\emph{Architectures for agents in regulated work.} Pattern catalogues and reference architectures for foundation-model agents provide bounded planning, guardrails, oversight and accountability~\cite{liu,lurai,lura}, and multi-layer runtime guardrails combine complementary controls~\cite{shams}; work on ML systems stresses checkable production controls~\cite{sculley,breck}. Closest in domain, Okpala et al.\ build agent crews for documentation compliance, model replication, conceptual soundness and outcome analysis~\cite{okpala}, and Wang et al.\ propose an SR~26-2-compatible control framework for generative AI in financial institutions~\cite{wang}. Neither specifies how to attribute an agent's contribution against a deterministic baseline, which is our focus.

\section{Architecture: What Is Built and What Is Proposed}\label{sec:arch}

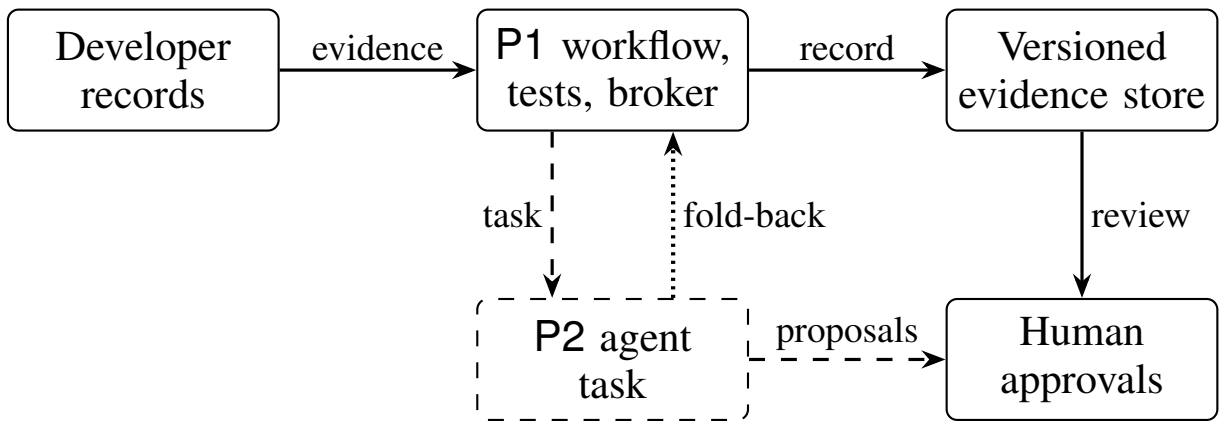
\begin{figure}[t]
\centering
\begin{tikzpicture}[font=\footnotesize, node distance=4mm and 13mm,
  box/.style={draw, rounded corners=2pt, align=center, minimum height=8mm, inner sep=2pt, text width=16.5mm},
  opt/.style={box, dashed},
  arr/.style={-{Stealth[length=1.6mm]}, semithick},
  darr/.style={arr, dashed},
  farr/.style={arr, densely dotted, thick},
  lab/.style={font=\scriptsize, inner sep=1.5pt}]
\node[box] (dev) {Developer\\records};
\node[box, right=of dev] (p1) {\Pone\ workflow,\\tests, broker};
\node[box, right=of p1] (ev) {Versioned\\evidence store};
\node[opt, below=11mm of p1] (agent) {\Ptwo\ agent\\task};
\node[box, below=11mm of ev] (hum) {Human\\approvals};
\draw[arr] (dev) -- node[lab, above]{evidence} (p1);
\draw[arr] (p1) -- node[lab, above]{record} (ev);
\draw[arr] (ev) -- node[lab, right]{review} (hum);
\draw[darr] ([xshift=-4mm]p1.south) -- node[lab, left]{task} ([xshift=-4mm]agent.north);
\draw[farr] ([xshift=4mm]agent.north) -- node[lab, right]{fold-back} ([xshift=4mm]p1.south);
\draw[darr] (agent) -- node[lab, above]{proposals} (hum);
\end{tikzpicture}
\caption{Trust boundary and attribution. Solid: \Pone\ runs without LLMs and is the control arm. Dashed: an optional \Ptwo\ task, credited with detection only for faults that the \Pone\ record does not report; it writes proposals, never approved decisions. Dotted: fold-back, through which an automatable agent finding becomes a \Pone\ check between rounds and earns a discovery credit. Developer records are evidence, not instructions. The broker, workflow and decision service are proposed components that the pack does not implement.}
\label{fig:arch}
\end{figure}

\Pzero\ is current validation practice. \Pone\ is a workflow with versioned evidence, mandatory deterministic tests, a fixed state machine and non-agent approval authority. \Ptwo\ adds one contract-bound agent task to the \emph{same} \Pone\ backbone, so removing the agent leaves \Pone\ intact (Fig.~\ref{fig:arch}). Developer records are evidence, never instructions. A task contract binds immutable record references, tools, an output schema, autonomy and a budget. Agents write proposed claims, requests and tests, never severity, approval or closure, and a failed write-back stays pending instead of becoming an approval. Table~\ref{tab:req} lists five requirements, which instantiate known patterns such as policy-driven workflows, guardrails and separation of duties~\cite{liu,lurai,shams}, together with what the pack implements.

\begin{table}[t]
\caption{Requirements, mechanisms and what the pack implements}\label{tab:req}
\centering\footnotesize
\begin{tabular}{@{}L{0.35cm}L{3.75cm}L{3.75cm}@{}}
\toprule
& Requirement and intended mechanism & Status in the pack \\
\midrule
R1 & Fixed state machine; agents cannot waive required work & \Pone\ coverage checks built; broker proposed \\
R2 & Reserved approvals; scoped tool access & Schema and contract authored; enforcement proposed \\
R3 & Typed, attributable claims; sources are evidence, not authority & Structural, neutrality and adjudication checks built \\
R4 & Version-bound evidence and approvals & Hashed records built; approval reconciliation proposed \\
R5 & Separate audit replay, numerical agreement and LLM stability & Deterministic replay built; agent stability untested \\
\bottomrule
\end{tabular}
\end{table}

The demonstration contract grants one read-only and one proposal-only tool and caps a run at 300\,s, 16 tool calls, one retry, US\$5, 32{,}000 tokens of context per call and 150{,}000 cumulative tokens, counting input and output over all calls including re-sent context. A static size audit, which is not a tokenizer measurement, puts peak context at 11--13 thousand proxy tokens (characters divided by four) and a sequential one-record-per-call read at 86--91 thousand cumulative input tokens before any system prompt or output. The dollar cap binds first only above US\$33.33 per million tokens. Nothing in the pack enforces the contract, and its feasibility is untested.

\section{Attribution Protocol}\label{sec:protocol}

\subsection{Comparison and attribution}
First compare \Pone\ with \Pzero\ to measure infrastructure gains. Then compare a frozen \Pone\ with \Pone\ plus \emph{one} agent task, holding inputs, deterministic tests, tools and review instructions fixed. Score the raw agent proposal before any human editing, and record later checking, rewriting and model-owner work as costs, not as agent correctness.

A seeded fault is \emph{\Pone-reported} if the \Pone\ record already exposes it and \emph{\Pone-unreported} otherwise; unreported faults are further split into currently automatable and contextual issues. An agent that repeats only \Pone-reported faults has a reported-fault miss rate of 0 and an unreported-fault miss rate of 1, yet a pooled miss rate of only 0.4 in this pack: pooling hides that it adds no detection. The protocol therefore names two value channels: \emph{detection beyond \Pone}, measured on unreported faults, and \emph{effort on reported faults} (root cause, materiality and drafting), measured only through net hours (Sec.~\ref{sec:gates}), never as detection.

\emph{Fold-back.} \Pone\ is frozen within a study round. Between rounds, any agent finding that an approved deterministic check could catch is promoted into \Pone, and the next round compares against the stronger baseline; this prevents work from being preserved artificially for the agent. Because promotion removes the finding from later increments, the agent whose finding motivated a promoted check receives a \emph{discovery credit}. Sec.~\ref{sec:policies} executes one round.

\subsection{Scoring rules}\label{sec:scoring}
Every decoy (a visible but harmless difference, such as an equivalent encoding) needs an explicit disposition, and every proposed request, test or escalation needs a necessity judgment, so over-caution has a visible cost. A developer assertion adopted anywhere in a response counts as endorsed, even if another paragraph challenges it. Structural inspection verifies the schema, identifiers and hashed evidence references but returns \texttt{semantic\_status: not\_assessed}; it never certifies prose. Two validators independent of authoring and delivery score every raw output, recording original ratings, Cohen's $\kappa$ where estimable and a third person's resolution of disputes. Primary rates are intention-to-treat: a missing, invalid or unreviewed output misses every fault of its assignment, and rates over the adjudicated subset are descriptive only. If the Q861 output in Sec.~\ref{sec:harness} were structurally invalid, for example, the subset miss rate for unreported faults would be 0.0, because only Q593 would be scored; the primary rate is 0.5.

\subsection{Mechanism tests, behavioral gates and economics}\label{sec:gates}
Enforceable invariants (permissions, approval separation, hash integrity and write-back recovery) belong at broker and service boundaries and must be checked by adversarial and fault-injection tests; observing obedient agents cannot establish them. Residual behavioral failures, such as unsupported claims, omissions and gratuitous escalation, need independent human evaluation and statistical gates. With zero failures in $n$ independent Bernoulli opportunities, the one-sided 95\% upper bound is $1-0.05^{1/n}$~\cite{hanley}: 12 independent families still allow about 22\%, and a 1\% bound needs 299. Variants of one family are clustered, not counted as independent families.

Let $C$ be the incremental fixed cost, $N$ the eligible engagements over a pre-registered horizon, $c$ the incremental non-labor cost per engagement and $r$ the cost per staff hour. Counting validators' checks, rework, escalations, adjudication and model-owner effort, mean \emph{net} hours saved per engagement must exceed
\begin{equation}
h^{*} = \frac{C/N + c}{r}. \label{eq:h}
\end{equation}
For $C=\$100{,}000$, $c=\$100$ and $r=\$150$, $h^{*}$ is 67.33, 34.00 and 14.00 hours for $N=10$, 20 and 50. The gate passes only if the lower endpoint of a two-sided 95\% paired confidence interval exceeds $h^{*}$. If one optional extension is pre-registered, both looks use a two-sided 97.5\% interval (conservative multiplicity control), not 95\% followed by 97.5\%~\cite{jennison}. Inconclusive results are not positive evidence.

\section{Synthetic Harness and Results}\label{sec:harness}

\subsection{Harness}
The releasable v0.7 pack contains six variants of one synthetic probability-of-default scorecard. Utilization $u\in[0,1]$ is binned at 0.3 and 0.8 with $\mathrm{WoE}=\ln[P(\mathrm{bin}\mid\mathrm{good})/P(\mathrm{bin}\mid\mathrm{bad})]$; then $z=-2.0-0.8\,\mathrm{WoE}$, $\mathrm{PD}=(1+e^{-z})^{-1}$ and $S=600+(20/\ln 2)\ln[(1-\mathrm{PD})/(50\,\mathrm{PD})]$, with approval at rounded $S\ge 550$. A Decimal reference and a manifest-driven NumPy float32 deployment are diverse implementations that share one author and specification, so common-mode mistakes remain possible~\cite{knight}. \Pone\ executes a specification-derived 40-input grid, covering boundaries, adjacent binary64 values, near-boundary probes and every invalid or missing input class, and it checks plan coverage. Every variant has the same twelve hashed, versioned records and one generic instruction.

Q718 deploys weight $+0.7$ instead of $-0.4$ in the high bin, turning a tested decline (score 536 at $u=0.9$) into an accept (561); \Pone\ reports 10 grid failures. In Q204 the deployed rules put exactly $u=0.8$ in the middle bin, understating PD by 4.60 percentage points and raising the score by 11.5 points. Both implementations still decline at the cutoff, so a decision-only comparison would pass; \Pone\ reports the failure at 0.8 and the plan's missing equality test. An agent that restates these items earns full credit on \Pone-reported faults but no detection credit. Its possible value lies in explaining the closure error and its materiality, which the protocol counts as hours. Nothing in \Pone's record flags Q593, whose developer note claims acceptable calibration while segment~B shows 45 observed against 20 expected defaults (tolerance 0.05), or Q861, whose feature study gives marginal district counts but no district-by-group or selection-impact evidence. Only these two faults can earn detection credit. Q437 is clean, and Q926 makes three valid, equivalent deployment changes that serve as decoys.

\subsection{Do the metrics separate known behaviours?}\label{sec:policies}
We scored five scripted reference policies on all six variants (Table~\ref{tab:policies}). \emph{Ideal} is the authored control response set. The other four are deterministic functions of a variant's agent-visible records: the \emph{parrot} restates what the \Pone\ record reports; \emph{escalate-all} adds a generic escalation and evidence request everywhere; the \emph{decoy flagger} also reports a reporting-only field and every equivalent change as defects; and \emph{anchored} adopts the developer's statements. Five fixed rules adjudicate the scripted outputs through the ordinary evaluator: (J1)~a claim restating a failed \Pone\ component is supported and matches each fault whose evidence cites that component, and a test proposed for a gap \Pone\ reports is necessary; (J2)~a generic escalation or evidence request that names no specific gap is unnecessary; (J3)~a claim reporting an equivalent encoding, an accurate statement or a reporting-only field as a defect is unsupported and consequential, and it flags each decoy whose evidence it cites; (J4)~adopting a developer statement endorses it, and is unsupported and consequential if the statement is incorrect; and (J5)~anything unmatched is missed or not mentioned. J1 deliberately credits restatement as detection, so reported-fault scores are upper bounds. Because the policies are scripted, the rules apply mechanically; for an agent, the same judgments come from the independent validators of Sec.~\ref{sec:scoring}. The policies and rules are authored, so this is a known-answer test of the metrics, not evidence about agents.

\begin{table}[t]
\caption{Scripted reference policies on the six variants (not agent results). Escal.: escalate-all; Anch.: anchored.}\label{tab:policies}
\centering\footnotesize
\setlength{\tabcolsep}{3.4pt}
\begin{tabular}{@{}lccccc@{}}
\toprule
 & Ideal & Parrot & Escal. & Flagger & Anch. \\
\midrule
\multicolumn{6}{@{}l}{\emph{Naive metrics, which ignore \Pone}}\\
Case accuracy & 6/6 & 4/6 & 4/6 & 4/6 & 4/6 \\
Pooled faults found & 5/5 & 3/5 & 3/5 & 3/5 & 3/5 \\
\midrule
\multicolumn{6}{@{}l}{\emph{Protocol metrics}}\\
\Pone-reported faults found & 3/3 & 3/3 & 3/3 & 3/3 & 3/3 \\
Increment: unreported found & 2/2 & 0/2 & 0/2 & 0/2 & 0/2 \\
Clean variants, false finding & 0/2 & 0/2 & 0/2 & 2/2 & 0/2 \\
Decoy false alarms & 0/14 & 0/14 & 0/14 & 9/14 & 0/14 \\
Unnecessary actions & 0 & 0 & 12 & 0 & 0 \\
Endorsed false assertions & 0/3 & 0/3 & 0/3 & 0/3 & 3/3 \\
Outputs with hard failure & 0/6 & 0/6 & 0/6 & 6/6 & 3/6 \\
\midrule
\multicolumn{6}{@{}l}{\emph{After one fold-back round (Q593 now \Pone-reported)}}\\
Case accuracy & 6/6 & 5/6 & 4/6 & 4/6 & 5/6 \\
Pooled faults found & 5/5 & 4/5 & 4/5 & 4/5 & 4/5 \\
Increment: unreported found & 1/1 & 0/1 & 0/1 & 0/1 & 0/1 \\
Discovery credits & 1 & 0 & 0 & 0 & 0 \\
\bottomrule
\end{tabular}
\end{table}

Naive metrics that ignore \Pone\ tie the four flawed policies at 4/6 case accuracy and 3/5 pooled faults found, and all three of those faults are \Pone's work. Under the protocol each policy has a distinct signature, and only the ideal policy adds detection beyond \Pone. Over-caution appears as 12 unnecessary actions; decoy flagging as 9 of 14 decoy false alarms and false findings on both clean variants; and anchoring as 3 of 3 endorsed incorrect assertions.

We then executed one fold-back round. A per-segment calibration check, recomputed from the outcome records, was promoted into \Pone\ and the pack was rebuilt; all 64 checks still pass, and Q593 becomes \Pone-reported. The parrot's naive scores rise to 5/6 and 4/5 without any change to the policy, while its increment stays at 0 of the one remaining unreported fault. The ideal policy's increment becomes 1/1, and it earns one discovery credit for Q593.

\subsection{Validity of the evaluation itself}
Earlier versions of the harness passed their own suites while containing nine defects (Table~\ref{tab:defects}): v0.3 passed 15 of 15 checks with defects 1--3, v0.4 passed 41 of 41 with defects 4--8, and v0.6 passed 62 of 62 with defect 9. Defects 1--8 were found during author-led, AI-assisted design reviews and defect 9 during an AI-assisted pre-submission review. Sixteen representative mutants that reconstruct these mechanisms are each rejected by named checks of the current suite, which has 7 integrity and 57 engineering checks. Three of the mutants survived until checks targeting them were added, so the matrix is regression evidence, not a measure of detection power. Likewise, 37 of the 57 engineering checks exercise the evaluator's bookkeeping on authored inputs, so their count describes coverage of the harness, not detection quality.

\begin{table}[t]
\caption{Nine ways earlier versions of the evaluation passed misleadingly. Each old suite passed all of its own checks (v0.3: 15, v0.4: 41, v0.6: 62); every mutant is rejected by the current suite.}\label{tab:defects}
\centering\footnotesize
\begin{tabular}{@{}L{0.15cm}L{5.5cm}L{0.6cm}L{1.0cm}@{}}
\toprule
 & Misleading pass mechanism & Suite & Mutants \\
\midrule
1 & Instruction, schema or record wording hints at the answer & v0.3 & M1a--d \\
2 & A structurally valid response is credited as correct; prose is never judged & v0.3 & M2 \\
3 & The deployment path reuses the reference code, so both share one oracle & v0.3 & M3 \\
4 & Faults \Pone\ already reports are credited to the agent: no \Pone\ record, or pooled rates only & v0.4 & M4a--b \\
5 & Decoy false alarms and unnecessary actions cost nothing & v0.4 & M5a--b \\
6 & Boundary and invalid-input gaps: a float32-binning deployment passed the old plan but fails 7 of the 40 current grid inputs & v0.4 & M6a--c \\
7 & The comparator never executes the candidate, so every comparison passes & v0.4 & M7 \\
8 & The verifier silently repairs a tampered case file & v0.4 & M8 \\
9 & Missing or invalid outputs leave the miss-rate denominators & v0.6 & M9 \\
\bottomrule
\end{tabular}
\end{table}

Lexical neutrality checks have a harder limit. Three paraphrased answer cues, such as a change log that mentions a ``superseded parameter table,'' each pass all 64 checks when rebuilt into a temporary copy of the pack. Semantic answer leakage therefore needs independent adversarial review and cannot be certified by forbidden-word lists, echoing known benchmark leakage failures~\cite{zhu,swebenchplus}.

\section{Limitations and Falsifiable Next Step}
The pack verifies its own deterministic mechanics and the behaviour of its metrics on scripted policies, not agent accuracy or the proposed architecture: it contains no agent runtime, enforced broker or independent reviewers. The reviews that found the nine defects were author-led with AI assistance, not independent audits. The six variants share one case family and may become training data once released, and the budget estimates are static character counts.

A prospective feasibility pilot will fix one model version and one submission-review task, use 12 independently authored private case families with two matched variants each (24 agent outputs), and blind two validators to arm and to each other. It will record real token and tool use, reviewer time, corrections, false alarms and every output, including incomplete ones, which count as misses. Feasibility fails if fewer than 20 of the 24 outputs are structurally complete, if validators cannot independently score more than 90\% of output units, if $\kappa<0.60$ for any estimable judgment type, or if median adjudication time exceeds 30 minutes per output. These thresholds are proposed, not achieved. The pilot cannot establish a 1\% safety rate or net economic value; a release decision would require substantially more independent evidence and separately verified engineering controls.

\section{Conclusion}
An LLM agent in model validation should be credited only for what a bank's deterministic workflow does not already do. Our architecture keeps that workflow as both fallback and control arm, and our protocol turns the comparison into explicit, falsifiable rules. On scripted policies, metrics that ignore the control arm cannot tell parroting, over-caution, decoy flagging and anchoring apart, while the protocol can. Whether any real agent adds anything remains open, and the pilot is designed to answer it.

\section*{Data Availability}
{\raggedright The synthetic evaluation pack, version 0.7, accompanies this preprint as the ancillary archive \texttt{eval-pack-v0.7-mit.zip}. Its \texttt{checks.py} runs 64 verification checks; \texttt{policies.py}, \texttt{audits/foldback.py} and \texttt{mutants/run\_mutants.py} reproduce Tables~\ref{tab:policies} and~\ref{tab:defects}; and \texttt{paper\_arithmetic.py} checks the numerical statements in the paper. Earlier harness versions are not included; their recorded results are provided in \texttt{mutants/results.json}. The SHA-256 digest of the pack's \texttt{SHA256SUMS.txt} manifest begins \texttt{c16e3186e04bf2a6}. The evaluation-pack code and synthetic fixtures are provided under the MIT License (see the pack's \texttt{LICENSE} file). All case data are synthetic; no proprietary bank data are included. The pack contains no evaluated LLM-agent outputs or implemented agent runtime.\par}

\section*{Acknowledgment: Generative-AI Use}
The research questions, study design, design decisions, and interpretation of results are the authors' own. Claude Opus 5.5 (Anthropic) was used to review the authors' experimental design, to generate portions of the evaluation harness code (including the scripted policies, the fold-back audit, and the intention-to-treat scoring), and to draft and edit text in all sections. ChatGPT GPT-6 (OpenAI) was used to restructure and reword an earlier draft, to convert the manuscript to \LaTeX, and to write supplemental audit scripts. The authors reviewed, tested, and revised all AI-generated text and code and take full responsibility for the content of this paper, including its accuracy, citations, and conclusions. All reported results were produced by running the authors' verified code; none were generated by an AI system.

\bibliographystyle{IEEEtran}
\bibliography{references}

\end{document}